%% file: main.tex
\documentclass[conference]{IEEEtran}
\IEEEoverridecommandlockouts

\usepackage{cite}
\usepackage[utf8]{inputenc} 
\usepackage[T1]{fontenc}    
\usepackage{hyperref}       
\usepackage{url}            
\usepackage{booktabs}       
\usepackage{amsfonts}       
\usepackage{nicefrac}       
\usepackage{microtype}      
\usepackage{xcolor}         
\usepackage{makecell}
\usepackage{colortbl}

\usepackage{booktabs} 
\usepackage{tikz} 
\usepackage{array}
\usepackage{amsmath}
\usepackage{amsthm}
\usepackage{amssymb}
\usepackage{graphicx}
\usepackage{subfigure}
\usepackage{xspace}
\usepackage{paralist}
\usepackage{multirow}
\usepackage{caption,latexsym}
\usepackage[ruled,vlined,oldcommands]{algorithm2e}
\usepackage{wrapfig}
\usepackage{url}
\usepackage{filecontents}
\usepackage{pgfplots}
\usepgfplotslibrary{fillbetween}

\newtheorem{definition}{Definition}
\newtheorem{example}{Example}

\newtheorem{thrm}{Theorem}
\newcommand{\eat}[1]{}

\begin{document}

\title{On the Verification of Embeddings with Hybrid Markov Logic
\thanks{This research was supported by NSF awards \#2008812 and \#1934745. The opinions, findings, and results are solely the authors' and do not reflect those of the funding agencies.}
}

\author{\IEEEauthorblockN{Anup Shakya}
\IEEEauthorblockA{\textit{University of Memphis} \\
ashakya@memphis.edu}
\and
\IEEEauthorblockN{Abisha Thapa Magar}
\IEEEauthorblockA{\textit{University of Memphis} \\
thpmagar@memphis.edu}
\and
\IEEEauthorblockN{Somdeb Sarkhel}
\IEEEauthorblockA{\textit{Adobe Research} \\
somdeb@adobe.com}
\and
\IEEEauthorblockN{Deepak Venugopal}
\IEEEauthorblockA{\textit{University of Memphis} \\
dvngopal@memphis.edu}
}

\maketitle

\input{abs}
\input{intro}
\input{related}
\input{verify}

\input{data}
\input{exps}
\input{conclusion}

\bibliography{main}
\bibliographystyle{IEEEtran}
\end{document}

%% file: abs.tex
\begin{abstract}
    The standard approach to verify representations learned by Deep Neural Networks is to use them in specific tasks such as classification or regression, and measure their performance based on accuracy in such tasks. However, in many cases, we would want to verify more complex properties of a learned representation. To do this, we propose a framework based on a probabilistic first-order language, namely, Hybrid Markov Logic Networks (HMLNs) where we specify properties over embeddings mixed with symbolic domain knowledge. We present an approach to learn parameters for the properties within this framework. Further, we develop a verification method to test embeddings in this framework by encoding this task as a Mixed Integer Linear Program for which we can leverage existing state-of-the-art solvers. We illustrate verification in Graph Neural Networks, Deep Knowledge Tracing and Intelligent Tutoring Systems to demonstrate the generality of our approach.
    
\end{abstract}

%% file: intro.tex
\section{Introduction}


A typical approach to verify Deep Neural Network (DNNs) representations (or embeddings) is to evaluate them in a specific downstream task (e.g. classification). To evaluate structure in an embedding, approaches such as~\cite{embeddingcomparator} visualize and compare the global structure of an embedding with a reference. While beneficial, such approaches are still fairly limited since they do not verify properties that combine the structure of embeddings with natural relationships that exist within the domain. For example, using the language of first-order logic, we can encode a property regarding the spread of flu as ${\tt Friends}(x,y)$ $\wedge$ ${\tt Dist}(x,y)<\tau$ $\Rightarrow$ $({\tt Flu}(x)\Leftrightarrow{\tt Flu}(y))$, where ${\tt Dist}(x,y)$ denotes the distance between embeddings learned by a DNN and $\tau$ is a threshold value. This property combines the geometry of the embedding (neighborhoods) with symbolic relations. In this paper, we develop a probabilistic framework to verify such properties.


Specifically, in our verification framework, we learn a {\em specification HMLN} (Hybrid Markov Logic Network~\cite{wang2008hmln}) that combine sub-symbolic representations with symbolic knowledge to represent verifiable properties.
We parameterize properties with weights that represent uncertainty and learn these weights from embeddings observed from a specification DNN. To account for noise that may be inherent in specification embeddings, and to reduce variance in the parameterization, we develop a Rao-Blackwellized likelihood function that we optimize to learn the specification HMLN. 
We then formulate verification of a property as probabilistic inference in the specification HMLN. Specifically, similar to software/hardware verification where we compare states of variables through assert statements, here, we compare the specification embedding with a test embedding by expressing bounds on the probability of a property conditioned on the embeddings. We show that we can compute these bounds by formulating a Mixed Integer Linear Program (MILP) which can be solved by existing state-of-the-art solvers. 

In our evaluation, we verify embeddings learned for varied applications illustrating the generalizability of our framework. Specifically, we i) verify semantic meaning of Graph Neural Networks embeddings and ii) verify invariance of embeddings learned through Deep Knowledge Tracing~\cite{piech2015deep} and iii) verify if embeddings can transfer across tasks in an Intelligent Tutoring System using data from MATHia, a commercial provider of K-12 Math learning software.




\eat{
To summarize, our main contributions are as follows:
\begin{itemize}
    \item A novel probabilistic framework for verification of (uncertain) properties over symbolic and sub-symbolic terms.
    \item Formalizing verification as a comparison between probability bounds on a property that can be computed leveraging existing state-of-the-art MILP solvers.
    \item Grounding our verification in real-world contexts including those domains that are critical to society such as education.
\end{itemize}
}

%% file: related.tex
\section{Related Work}


An overview of formal verification methods for DNNs is provided in \cite{albarghouthi-book}. Verification methods may be {\em complete} with stronger guarantees at the cost of poor scalability or {\em incomplete} that are more scalable with reduced guarantees~\cite{batten2021verification}. 
Deductive verification methods in \cite{Katz19Marabou} use constraint solvers to verify DNN properties.
In \cite{ijcai2022p503}, a Neuro-Symbolic approach is proposed for verification which allows for richer specifications. 
Other types of approaches used for verification include abstract interpretation where the neural network layers are converted to process on abstract inputs and produce abstract outputs that over-approximates the behavior of the neural network over all real inputs~\cite{NEURIPS2019_5df0385c, gagandeep2019}. For verifying local robustness, \cite{batten2021verification} proposed bounds of verification through chordal graph decomposition. MILP formulations have also been used to verify robustness in \cite{bastani2016measuring, Tjeng19mip, Botoeva2020}. However, in our case, the semantics of HMLNs allow us to specify more expressive relational properties combining symbolic and sub-symbolic representations.


%% file: verify.tex
\section{Specification Language}

To verify a model, we require a language to express properties for a model. First-order logic (FOL) is arguably the de-facto representation language that has been widely used for formal verification. In particular, constructs in FOL can be used to express symbolic and numeric terms and thus, we can represent complex properties of the model under verification by relational formulas containing discrete and continuous variables. We formalize this with the following definitions.


\begin{definition}
    A sub-symbolic atom is a numeric FOL term where the domain of all variables in the atom are sub-symbolic representations. A symbolic atom is a FOL term where the domain of all variables in the atom are constants.
\end{definition}

\begin{definition}
    A verifiable property is a FOL formula with terms that include both symbolic and/or sub-symbolic atoms. A ground property is one where all variables in the property have been replaced by an element from their respective domains.
\end{definition}

\begin{example}
${\tt Word}(w_1,x)$ $\wedge$ ${\tt Word}(w_2,y)$ $\Rightarrow$ ${\tt SameTopic}(x,y)$ is a verifiable property that contains only symbolic terms. $(D(e_{w},e_{w'})<\tau)$ * $({\tt Word}(w,x)$ $\wedge$ ${\tt Word}(w',x)$ $\Rightarrow$ ${\tt Sentiment}(x,s))$ is a property where $D(e_w,e_{w'})<\tau$ is sub-symbolic term that represents a continuous function over sub-symbolic representations $e_w,e_{w'}$ (e.g. distance between embeddings). ${\tt Word}(W_1,X)$ $\wedge$ ${\tt Word}(W_2,Y)$ $\Rightarrow$ ${\tt SameTopic}(X,Y)$ is a ground property where $W_1,W_2,X,Y$ are elements from the domains of the variables $w_1,w_2,x,y$ respectively.
\end{example}


\begin{definition}
    A set of verifiable properties are parameterized as a Hybrid Markov Logic Network (HMLN) distribution where i) each ground symbolic or sub-symbolic atom is a node in the Markov network and ii) each ground property is a feature in the Markov network.
\end{definition}

The probability distribution of the HMLN follows a log-linear model. Specifically, given $\{f_i;w_i\}_{i=1}^k$ where $f_i$ is a property and $w_i$ is a real-valued weight, the distribution is given by,
\begin{equation}
\label{eq:prob}
    P({\bf X}={\bf x}) = \frac{1}{Z}\exp\left(\sum_i w_is_i({\bf x})\right)
\end{equation}
where ${\bf x}$ is a {\em world}, i.e., an assignment to all ground atoms in the HMLN and $s_i({\bf x})$ is the value of the $i$-th formula. If $f_i$ is a property with only symbolic atoms, then $s_i({\bf x})$ is equal to the number of true groundings of $f_i$ given the assignment ${\bf x}$ and if $f_i$ is a property with continuous terms, $s_i({\bf x})$ is the value of the property $f_i$ given assignment ${\bf x}$. 


To express properties over sub-symbolic atoms, we utilize continuous functions. While the representation is capable of handling generic continuous functions, in typical verification tasks, the most common functions are equality and inequality. These functions are represented in the HMLN distribution as {\em soft} equality and inequality. Specifically, the soft equality value is equal to the square of the difference between numeric terms involved in the equality. The soft inequality value is equal to the negative log-sigmoid function~\cite{wang2008hmln}. For instance, $D(X,Y)==D(Y,X)$ is a soft equality, where the value of this term is given by $-(D(X,Y)-D(Y,X))^2$. Similarly, $D(X,Y) < \tau$ is a soft-inequality, where the value of the term is equal $-log(1+exp(a(\tau-D(X,Y))))$ and for $D(X,Y) > \tau$, the value is equal to $-log(1+exp(a(D(X,Y)-\tau)))$, where $\tau$ is the threshold and $a$ denotes the softness of the sigmoid.

\section{Specification Learning}

We learn a specification HMLN by parameterizing the properties $\{f_i\}_{i=1}^k$ with weights ${\bf w}$ $=$ $\{w_i\}_{i=1}^k$ based on an {\em observed world}. An observed world is an assignment to all ground atoms in the HMLN.
Specifically, let ${\bf X}$ represent all ground atoms and ${\bf X}={\bf x}$ denote an observed world, i.e., an assignment to ${\bf X}$. While in a typical case, we can learn the parameters by maximizing the log-likelihood of the observed world, here, note that the observed world contains both symbolic and sub-symbolic atoms. The assignments to symbolic atoms are typically derived from domain knowledge while for sub-symbolic atoms, the values come from a specification DNN and these values may be a source of uncertainty in learning the model.

Specifically, let ${\bf w}^*$ be the optimal weights for verifying properties $\{f_i\}_{i=1}^k$. Let $\hat{\bf w}$ be the weights learned by optimizing the log-likelihood. Let $\sigma^2(\hat{\bf w})$ denote the variance in the estimated weights. To reduce $\sigma^2(\hat{\bf w})$, we {\em Rao-Blackwellize} the log-likelihood by summing out sub-symbolic atoms from the function. Specifically, the Rao-Blackwellized objective is,
\begin{align}
\label{eq:rbll}
    \ell^{RB}({\bf w},{\bf y},{\bf x}_s,{\bf x}_e) = 
    \log P_{\bf w}^{RB}({\bf Y}={\bf y}|{\bf X}_e) = \nonumber\\ 
    \log \sum_{{\bf x}_s}P_{\bf w}({\bf y},{\bf x}_s|{\bf X}_{e})
\end{align}
where ${\bf X}_{e}\subseteq{\bf X}$ denote {\em evidence atoms}, i.e., symbolic atoms whose assignments are assumed to be known during inference, ${\bf X}_s\subseteq{\bf X}$ denote sub-symbolic atoms and ${\bf Y}={\bf X}\setminus ({\bf X}_{e}\cup {\bf X}_{s})$ denote {\em query atoms}, namely, atoms whose assignments are unknown and need to be predicted during inference. The partial derivative for the Rao-Blackwellized objective w.r.t the weights is as follows.

\eat{
\begin{align}
\frac{\partial(\ell^{RB}({\bf w},{\bf y},{\bf x}_s,{\bf x}_e))}{\partial{w_i}} = 
&\sum_{{\bf x}_s}P_{\bf w}({\bf y},{\bf x}_s|{\bf x}_e)*s_i({\bf y},{\bf x}_s,{\bf X}_e) -\nonumber\\
&\sum_{{\bf y},{\bf x}_s}P_{\bf w}({\bf y},{\bf x}_s|{\bf X}_e)*s_i({\bf y},{\bf x}_s,{\bf x}_e)
\end{align}
We can write the above equation in terms of the difference between expectations as follows.
}
\begin{align}
\label{eq:rbl}
\frac{\partial(\ell^{RB}({\bf w},{\bf y},{\bf x}_s,{\bf x}_e))}{\partial{w_i}} =
&\mathbb{E}_{{\bf w},{\bf x}_s}[s_i({\bf y},{\bf x}_s,{\bf X}_e)] - \nonumber\\
&\mathbb{E}_{{\bf w},{\bf x}_s,{\bf x}_e}[s_i({\bf y},{\bf x}_s,{\bf x}_e)]
\end{align}
where $\mathbb{E}_{{\bf w},{\bf x}_s}$ denotes that we compute the expectation w.r.t $\sum_{{\bf x}_s}P({\bf y}|{\bf X}_e)$ and $\mathbb{E}_{{\bf w},{\bf x}_s,{\bf x}_e}$ denotes that the expectation is w.r.t $\sum_{{\bf x}_s,{\bf y}}P({\bf y}|{\bf X}_e)$. Using this, we can directly apply gradient descent to update the weights.
Note that in each gradient computation, we need to estimate two expectations. Since computing them exactly is intractable, we use approximate inference to estimate each of them.  Let $\hat{\bf w}^*$ denote the converged weights when we use an unbiased (or asymptotically unbiased) estimator for both $\mathbb{E}_{{\bf w},{\bf x}_s}$ and $\mathbb{E}_{{\bf w},{\bf x}_s,{\bf x}_e}$. From the Rao-Blackwell theorem, it follows that $\sigma^2(\hat{\bf w}^*)$ $\leq$ $\sigma^2(\hat{\bf w})$. 

\subsection{MILP Encoding}

Similar to the Voted Perceptron for MLNs~\cite{domingos&lowd09}, we use MAP inference to estimate the intractable expectations. However, in our case, since we have both discrete and continuous variables in the HMLN, we formulate MAP inference as a Mixed Integer Linear Program (MILP). With this formulation, we can leverage state-of-the-art solvers to scale up weight learning even when the number of ground properties are very large.
Specifically, the MILP objective is as follows.
\begin{equation}
\label{eq:mobj}
 \arg\max_{{\bf x}}\exp\left(\sum_i w_is_i({\bf x})\right)\approx  \arg\max_{{\bf x}}\sum_i w_is_i({\bf x}) 
\end{equation}
Once we obtain the solution to the MILP, we compute the value of the $i$-th formula in the solution as an approximation of its expected value in the gradient computation. Since we need two expected values, we run the MILP twice. In the first case, only the evidence atoms are assumed to be known, i.e., we add constraints that assign evidence atoms to their respective values in the observed world. Thus, the MILP solution computes assignments jointly over both the sub-symbolic and query atoms. In the second case, we assume that both the evidence and sub-symbolic atoms are known, i.e., we add constraints to assign these to their respective values and the MILP computes the assignments to only the query atoms.

We encode a property ($w_i,f_i$) in the MILP as follows.
\eat{
$M(f_i)$ = \begin{cases}
  $f_i\Leftrightarrow a_i$  & \text{if $f_i$ is symbolic}\\
  $f_i\Leftrightarrow a_i$ & n \text{ is odd}
\end{cases}
}
If $f_i$ only contains symbolic atoms, we encode an auxiliary variable $a_i$ whose objective value is $w_i$ and an equivalence relation $f_i\Leftrightarrow a_i$ as a linear constraint. If $f_i$ contains both symbolic and sub-symbolic atoms, we encode each sub-symbolic atom in $f_i$ as a continuous variable, say $a_{i1}^c$ $\ldots$ $a_{ik}^c$.
For the symbolic sub-formula within $f_i$, say $f_{is}$, we encode a binary auxiliary variable $a_i$ and constraint $f_{is}\Leftrightarrow a_{i}$. The objective value for $a_i$ is $w_i*g(a_{i1}^c \ldots a_{ik}^c)$, where $g()$ encodes the value of $f_i$. Note that the value of $f_i$ may be a non-linear expression over $a_{i1}^c \ldots a_{ik}^c$. For instance, both the soft equality and inequality functions over sub-symbolic variables are non-linear functions. In such cases, we perform a relaxation by adding constraints to encode $g()$ as a piece-wise linear approximation to the true value of $f_i$.








\subsection{Weight Sharing}

Thus far, we assumed that each property is parameterized by a single weight. However, in practice, we may need multiple weights for the same property. For instance, suppose we are verifying an embedding learned by a deep neural network model; the geometry, sparsity, and density of points can vary across the embedding space. Thus, we may need multiple weights for the same property when defined over this space. To do this, we {\em share} weights over subsets of property groundings defined as hypercubes~\cite{Singla_Nath_Domingos_2014}.

\begin{definition}
A hypercube is a vector ${\bf H}$ $=$ $[{\bf S}_1 \ldots {\bf S}_m]$ where each ${\bf S}_i$ is a set of constants from a domain.
\end{definition}


\begin{definition}
The projection of a hypercube ${\bf H}$ $=$ $[{\bf S}_1 \ldots {\bf S}_m]$ on a property $f$ is the set of possible groundings of $f$ where each variable in $f$ is substituted by $S$ $\in$ ${\bf S}_1 \times \ldots \times {\bf S}_m$.
\end{definition}

\begin{example}
    Suppose we have three domains $\Delta_x$ $=$ $\{X_1,X_2,X_3\}$, $\Delta_y$ $=$ $\{Y_1,Y_2,Y_3\}$, $\Delta_z$ $=$ $\{Z_1,Z_2,Z_3\}$, the projection of the hypercube $[\{X_1,X_2\},\{Y_1,Y_2\},\{Z_1\}]$ on ${\tt R}(x)\wedge{\tt S}(x,y)$ is a set of 4 ground properties. The projection of the same hypercube on ${\tt T}(z)\wedge{\tt S}(z,y)$ is a set of 2 ground properties.
\end{example}

\begin{definition}
Given two hypercubes ${\bf H}_1$, ${\bf H}_2$, the merge operator ${\bf H}_1$ $\oplus$ ${\bf H}_2$ generates a hypercube that contains all sets in ${\bf H}_1$ that have no elements in common with any set in ${\bf H}_2$, all sets in ${\bf H}_2$ that have no elements in common with any set in ${\bf H}_1$ and for two sets ${\bf S}$ $\in$ ${\bf H}_1$, ${\bf S}'$ in ${\bf H}_2$ where ${\bf S}\cap {\bf S}'\not = \emptyset$, ${\bf H}_1 \oplus {\bf H}_2$ contains ${\bf S}\cap {\bf S}'$.
\end{definition}

We recursively construct a set of disjoint hypercubes. First, we define a bounding hypercube that consists of all domains.  In each recursive step, we select a hypercube ${\bf H}$ and split it into two disjoint hypercubes ${\bf H}^+$ and ${\bf H}^-$. To do this, we split a set ${\bf S}$ in ${\bf H}$ such that the average value of groundings in ${\bf H}^+$ is greater than in ${\bf H}$ and the average value of groundings in ${\bf H}^-$ is smaller than in ${\bf H}$. Since it is infeasible to evaluate all subsets in ${\bf S}$, we instead split ${\bf H}$ into $\{{\bf H}_i\}_{i=1}^k$, corresponding to $k$ constants in ${\bf S}$. We then merge all hypercubes into ${\bf H}^+$ where the average value exceeds that in ${\bf H}$ and similarly merge all hypercubes into ${\bf H}^+$ where the average value is smaller than in ${\bf H}$.

Given a set of disjoint hypercubes, $\{{\bf H}_i\}_{i=1}^k$, for each property $f$, we project each ${\bf H}_i$ on $f$ and learn a shared weight for all groundings in the projection.
To formally analyze the effect of weight sharing, let $w^*_1$ $\ldots$ $w^*_m$ denote the optimal parameterization, i.e., the weights that are required to verify the properties encoded in the HMLN. Suppose we instead use $k$ hypercubes and the total number of properties is $n$, we learn a total of $k*n$ weights. We now define a function $Q()$ that maps an optimal weight into one of the $k*n$ learned weights so as to minimize the absolute difference across all weights. Let $max_i |w^*_i-w_i|$ $\leq$ $\epsilon$, where $\epsilon$ is a constant. 
Similar to the results derived for quantization of weights in probabilistic graphical models~\cite{Carkhel_Gogate_2018}, we have the following.

\begin{thrm}
    $\ell({\bf w}^*,{\bf x}) - \ell({\bf w},{\bf x})$ $\leq$ $2kn\epsilon$
\end{thrm}

\eat{
\begin{proof}
    $$\ell({\bf w}^*,{\bf x}) = \sum_i w_i^*s_i({\bf x}) - \log\left(\sum_{{\bf x}'}\exp\left(\sum_iw_i^*s_i({\bf x}')\right)\right)$$
    $$\leq \sum_i (w_i+\epsilon)s_i({\bf x}) - \log\left(\sum_{{\bf x}'}\exp\left(\sum_i(w_i-\epsilon)s_i({\bf x}')\right)\right)$$
    \begin{align}
      \sum_i\epsilon s_i({\bf x})+\sum_iw_is_i({\bf x})-
      &\log\left(\sum_{{\bf x}'}\left(\exp(\sum_i(w_i)s_i({\bf x}')\right)\right)\nonumber\\
      &- \log\left(\exp\left(\sum_{i}\epsilon s_i({\bf x}')\right)\right)\nonumber  
    \end{align}

    If we assume that the value of each ground property is bounded (between 0 and 1), then we get the following inequality.
    \begin{align}
    &\leq \epsilon*k*n + \sum_iw_is_i({\bf x}) -
    \log\left(\sum_{{\bf x}'}\exp\left(\sum_i(w_i)s_i({\bf x}')\right)\right)\nonumber\\
    &+ \epsilon*k*n\nonumber\\
    &\leq 2\epsilon*k*n+\ell({\bf w},{\bf x})\nonumber
    \end{align}
    
\end{proof}
}

Thus, as we increase the number of hypercubes $k$, we can tighten the bound but at the same time, it increases the number of learnable weights. Thus, we need to achieve a trade-off between the two. In practice, given a bound on the number of hypercubes $\alpha$, we continue refining the hypercubes until the number of hypercubes $k\leq\alpha$.

\section{Verification}

In assertion-based verification in software/hardware systems, we describe valid/invalid behavior in the form of executable assert statements where we compare variables  to known valid/invalid states. Analogously, here, we compare embeddings based on properties in the specification HMLN. Specifically, we compare the embeddings generated from a specification DNN to those from a test DNN w.r.t a property. 
\eat{
One possible approach to perform this comparison is to compute the divergence between the probability distributions for a property conditioned on the test and specification embeddings. However, as the complexity of the property increases, it becomes harder to estimate probabilities. Specifically, Gibbs sampling~\cite{geman1984stochastic}, a widely-used sampling method has a large {\em mixing time} when the properties are complex. Specifically, the mixing time is dependent upon the {\em total influence} of an atom~\cite{pmlr-v48-sa16}. That is, if a change to the state of an atom affects a large number of ground formulas, the mixing time becomes infeasibly large. Further, when the variables are correlated (as is often the case in first-order probabilistic models), Gibbs sampling tends to get stuck in local optima and fails to explore the sampling space. Other samplers such as MC-SAT~\cite{poon&domingos2006} handle correlated variables more effectively. However, MC-SAT is a slice-sampler, and obtaining a slice of the distribution is equivalent to obtaining uniformly random samples from solutions to a SAT problem (and in our case a Satisfiability Modulo Theory due to continuous variables) which is again a hard task. At the same time, using the MILP encoding described in the prior section, we can perform MAP inference efficiently using existing solvers. We next describe our approach to perform verification using MAP values.
}
To illustrate what a prototypical verification may look like, we start with an example of a commonly occurring property of embeddings. Specifically, suppose we are given a specification DNN embedding ${\bf E}^*$ and want to verify if the embedding $\hat{\bf E}$ is similar, we can specify this as the following property.
\begin{equation}
    f=(D(x,y)< \tau)*[{\tt Class}(x)\Leftrightarrow {\tt Class}(y)]
\end{equation}
where $D(x,y)$ is the distance between embeddings of $x$ and $y$, and $\tau$ is a threshold. The verification statement for this is as follows.
\begin{equation}
    |\Omega_U(f|{\bf E}^*)-\Omega_U(f|\hat{\bf E})| \leq \delta_1 \wedge |\Omega_L(f|{\bf E}^*)-\Omega_L(f|\hat{\bf E})| \leq \delta_2
\end{equation}
where $\delta_1$ and $\delta_2$ are constants, $\Omega_U$ is an upper bound on the probability of the property in the HMLN distribution and $\Omega_L$ is a lower bound. To generalize, in order to verify a property $f$ for a test embedding based on a specification, we compute the absolute difference between bounds on the probability distribution of $f$ in the HMLN when conditioned on these embeddings. Next, we show that we can estimate these bounds through MAP values.

We consider a special case which we term as {\em constrained evidence set}. Specifically, $\{{\bf E}_1,\ldots{\bf E}_k\}$ is a constrained evidence set if the evidence on symbolic atoms remain fixed in all ${\bf E}_i$ but evidence on the sub-symbolic atoms are allowed to vary. This models a typical situation when we are trying to verify sub-symbolic representations learned through different methods while domain knowledge which is used to specify evidence on the symbolic atoms remains fixed. Note that by conditioning an HMLN on any ${\bf E}_i$, we condition on all the sub-symbolic atoms and thus, we obtain a discrete conditional distribution.

Given a HMLN and constrained evidence set $\{{\bf E}^*$, $\hat{\bf E}\}$, for a ground property $f$ in the HMLN, let $f_s=1$ indicate that the symbolic sub-formula in $f$ is true and $f_s=0$ that it is false. Let $(M^*_+,M^*_-)$ be the MAP values in the distribution conditioned on $\{{\bf E}^*,f_s=1\}$ and $\{{\bf E}^*,f_s=0\}$ respectively. Similarly, let $(\hat{M}_+,\hat{M}_-)$ be the MAP values in the distribution conditioned on $\{\hat{\bf E},f_s=1\}$ and $\{\hat{\bf E},f_s=0\}$ respectively. Using the log sum of exponential bounds, we can show the following result (we skip the proof for lack of space).


\begin{thrm}
$|\Omega_U(f|{\bf E}^*)-\Omega_U(f|\hat{\bf E})|$ $\propto$ $|(M^*_+-M^*_-)-(\hat{M}_+-\hat{M}_-)|$ and $|\Omega_L(f|{\bf E}^*)-\Omega_L(f|\hat{\bf E})|$ $\propto$ $|\hat{M}_--M^*_-|$.
\end{thrm}

In the above theorem, we assumed that the property is a ground property. To lift verification to a first-order property $f$, we compute the difference in mean values of the bounds over all groundings of $f$. Specifically, let $\mu_u$ be the mean difference between the upper bounds of the specification and the test embedding, and $\mu_l$ be the mean difference between the lower bounds.
Since the number of groundings may be very large, we  estimate $\mu_u$ and $\mu_l$ from sampled groundings of $f$.
Let $\{U_i^*\}_{i=1}^k$ and $\{L_i^*\}_{i=1}^k$ represent the upper and lower bounds for $k$ sampled groundings of $f$ using the specification, and $\{\hat{U}_i\}_{i=1}^k$ and $\{\hat{L}_i\}_{i=1}^k$ represent the bounds for the test embedding. 
The sample mean differences are computed as $\hat{\mu}_u$ $=$ $\frac{1}{k}\sum_{i=1}^k|U_i^*-\hat{U}_i|$ and $\hat{\mu}_l$ $=$ $\frac{1}{k}\sum_i|L_i^*-\hat{L}_i|$. Since the variances of the bounds computed from the test and specification embeddings may be different, we use {\em Welch's T-Test}~\cite{welch1947} to estimate $\mu_u$ and $\mu_l$ from $\hat{\mu}_u$ and $\hat{\mu}_l$. Thus, with a confidence interval of $1-\gamma$, we can verify the difference between the bounds computed from the test and specification embeddings.
Note that if $f$ contains a large number of groundings, the variance in the estimated difference of means could be large if we uniformly sample the groundings of $f$. Therefore, to reduce variance, we sample a grounding of $f$ from each hypercube and perform verification using these samples. Recall that within each hypercube all groundings of $f$ share the same formula weight. Thus, we cover all uniquely-weighted groundings of $f$ in the verification and at the same time, the computational complexity of verification is bounded by the number of hypercubes.

%% file: exps.tex
\section{Experiments}

\eat{
Finally, we verify embeddings learned from real-world student data obtained from a commercial  Intelligent Tutoring System (ITS) that provides one-on-one instruction to students in K-12 Math problem-solving. 
}

\subsection{Setup}
\eat{
We use Gurobi to solve the MILP during learning and inference. We set the hyper-parameters for learning the HMLN using grid-search, namely, learning rate of $0.01$ for gradient descent and the weight sharing parameter $\alpha$ to set the upper limit on the number of hypercubes to 200. We consider confidence intervals $\gamma$ for verification as $0.05$. We train the DNN models and the HMLN on a machine with 64 GB RAM and a Tesla GPU. 
}

We utilize Gurobi to solve the MILP problem both during the learning phase and during inference. To determine the optimal configuration for training the HMLN, we perform a grid search over hyper-parameters. Specifically, we set the learning rate for gradient descent to 0.01 and set the upper limit to number of hypercubes as 200. For the verification process, we consider a confidence interval, $\gamma$ as $0.05$. We refer to the DNN we are verifying as the {\em Network Under Verification} (NUV). Our code is available here\footnote{https://github.com/anupshakya07/verification}.

\subsection{GNN Verification}

To verify GNNs, we set up the following experiment. We learn node embeddings from three well-known GNN architectures, Graph Convolution Networks (GCNs)~\cite{kipf2017semisupervised}, Graph Attention Networks (GATs)~\cite{gat2018}, and GraphSage (GS)~\cite{hamilton2018inductive} with node classification as the downstream task. We use three standard benchmark datasets, namely, Cora, Citeseer, and PubMed for these experiments. 




\input{hmlns}

\input{results}

For designing the specification DNN, we use an approach similar to cross-validation, namely, we consider a DNN architecture as the specification to verify the remaining DNNs. We use the abbreviations GCN-S, GAT-S, and GS-S to denote the DNN specification.
Our specification HMLN for this task is the same for all three datasets and is shown in Table.~\ref{tab:hmlns} (with $\tau=0.5$). 
Here, we encode the {\em homophily property} over neighbors and a property relating distances between embeddings to the classes of those nodes. Thus, a closer alignment between the semantic information encoded in the embeddings and the symbolic knowledge implies that the verifier has greater confidence in the embeddings.



Table~\ref{tab:gnn-results} shows our results for all GNN verification tests. The values in red indicate {\em failed verification}. That is, when the Welch T-Test has $p\leq 0.05$ indicating that the NUV bounds (UB/LB) deviate from the specification. As seen here, for the original benchmarks, the verification passes for the benchmarks for most specifications which implies that the embeddings indeed encode semantic meaning.
We then created alternate benchmarks by perturbing the original graphs in an approach similar to \cite{pmlr-v206-alchihabi23a}, i.e., by adding noise to the graphs such that accuracy remains approximately the same, i.e., classification accuracy cannot be used for verification. 
On the other hand, the verification results on the noisy benchmarks in Table~\ref{tab:gnn-results} show that the semantic meaning encoded in the embeddings is significantly altered in several cases. In some cases, the meaning is preserved even in the presence of noise. For instance, when the specification is GCN and the NUV is GAT or vice-versa. One possible explanation could be that the aggregation of local neighborhoods using the attention mechanism helps in minimizing the change in embedding meaning in the presence of noise. On the other hand, GS though more scalable, since it approximates the local neighborhoods through samples, could result in a loss of semantic information in the embeddings.

\subsection{Knowledge Tracing Verification}

Deep Knowledge Tracing (DKT)\cite{piech2015deep} utilizes a DNN to learn embeddings representing student knowledge which helps us predict their future performance. To illustrate verification in this model, we use publicly available datasets provided in \cite{piech2015deep} which uses Item Response Theory (IRT)~\cite{irt1990} to generate the student data.
\begin{table}
    \centering
    \caption{Verification of Knowledge Tracing embeddings testing invariance to problem exchanges. $Student-p-n-c$ to denote that there are $p$ problems, $n*1000$ students and $c$ latent concepts. The t-statistic is shown in each case and the ones marked in red are those where verification fails for the NUV, i.e., the p-value $\leq$ $0.05$ for the Welch T-Test.}
    \resizebox{0.49\textwidth}{!}{
    \tabcolsep=0.15cm
    \begin{tabular}{|c|c|c|c|c|c|c|}
        \hline
        \multirow{2}{*}{\textbf{Dataset}}& \multicolumn{2}{|c|}{\textbf{DKT-E}} & \multicolumn{2}{|c|}{\textbf{DKT-H}} & \multicolumn{2}{|c|}{\textbf{DKT-P}}\\
        \cline{2-7}
        & \textbf{UB} & \textbf{LB} & \textbf{UB} & \textbf{LB}& \textbf{UB} & \textbf{LB}\\ 
         \hline
         \hline
         \texttt{Student-50-1-2} & -0.288 & -0.117 & \textcolor{red}{0.918} & 0.86 & 1.4 & 1.437\\  
         \hline
         \texttt{Student-100-4-5} & 0.991 & 1.875 & \textcolor{red}{2.38} & 1.7 & 1.95 & 1.671\\ 
         \hline
         \texttt{Student-50-4-5} & 0.561 & 1.12 & \textcolor{red}{2.13} & \textcolor{red}{3.51} & 0.981 & 1.163\\ 
         \hline
    \end{tabular}
    }
    \label{tab:student_result}
\end{table}
Specifically, in IRT, each problem is related to a latent concept, and the problems are of varying difficulty levels. The idea is to sample the responses of a student based on their skills and problem difficulty. The students' skills in a concept improve each time they encounter a problem from the same concept. We label the datasets as follows, $Student-p-n-c$ to denote that there are $p$ problems, $n*1000$ students and $c$ latent concepts. In this task, we verify {\em exchangeability} in DKT models. Specifically, the ordering of problems plays an important role in KT. Our goal in this task is to verify if DKT embeddings are {\em invariant to ordering of problems}.

The specification DNN is the DKT model developed in~\cite{piech2015deep} where an LSTM is trained to predict the next response of a student to a problem based on a sequence of student responses. The specification HMLN is shown in Table~\ref{tab:hmlns}. The first property specifies that a problem can be solved based on their pre-requisite knowledge. Specifically, for problems $p_1,p_2$ we consider $p_1$ as a pre-requisite of $p_2$ if they are both from the same latent concept and $p_1$ is easier than $p_2$.
In the next property, we encode that two students who have similar knowledge embeddings have similar capabilities in solving a problem.

We design the NUVs to verify if the DKT model learns similar embeddings when the ordering of problems that students work on have constraints. Specifically, in DKT-E, we learn an exchangeable LSTM using the idea in ~\cite{piech2015deep} where we train the model over several different orderings of the problems. In DKT-P, we constrain the ordering where, given the original ordering, we only exchange problems while preserving the prerequisite structure. Specifically, we exchange two problems $p_1,p_2$ from the same concept if $p_2$ is easier than $p_1$, i.e., students learn to solve progressively harder problems. In DKT-H, we do the opposite, i.e., exchange problems in the original ordering if they are from the same concept such that harder problems appear before easier problems in the ordering.

Table~\ref{tab:student_result} presents the verification results on three datasets. As seen here, for DKT-E, the verification passes for all datasets since the exchangeable model learns from different orderings. For DKT-P, the embeddings remain similar to the specification embeddings as indicated by the verification results. For DKT-H, there is a shift in the knowledge of students since it violates the prerequisite structure as shown by our verification results.



\subsection{ITS Verification}




Intelligent Tutoring Systems (ITSs) provide feedback to students during one-on-one interactions. In this task, we verify how well embeddings can {\em transfer} from learning a student's ability to providing adequate feedback which can help an ITS adapt to the student's learning. We use the Carnegie Learning MATHia 2019-20 dataset containing interactions between students and the ITS, publicly available through DataShop~\cite{StamperKBSLDYS11a}. Our specification is a transformer model used in~\cite{shakya_23} that predicts if a student gets a step in a problem correct or not. The NUV is a DNN that predicts the level of hints for each step (level-0 being minimal hints and level-3 indicating maximum hints). Our goal is to verify if embeddings learned in the specification can {\em transfer} to the NUV task.

In the HMLN shown in Table~\ref{tab:hmlns}, we define a common predicate called ${\tt Success}(s,p)$ that defines the criteria for success for a student in both the specification and the NUV DNNs. In the specification, we define ${\tt Success}(S,P)$ $=$ $1$ if the student $S$ gets more than $75\%$ of the steps correct in problem $P$. In the NUV, we define ${\tt Success}(S,P)$ $=$ $1$ if the student $S$ uses less than $25\%$ hints for problem $P$ that are level-2 or above. The specification HMLN encodes a property that if a student is successful in a problem, then they are likely to be successful in other problems from the same topic. Further, we also define a property where the similarity in embeddings in the specification DNN transfers to the similarity of success in the NUV.

\begin{figure}
    \centering
    \includegraphics[scale=0.48]{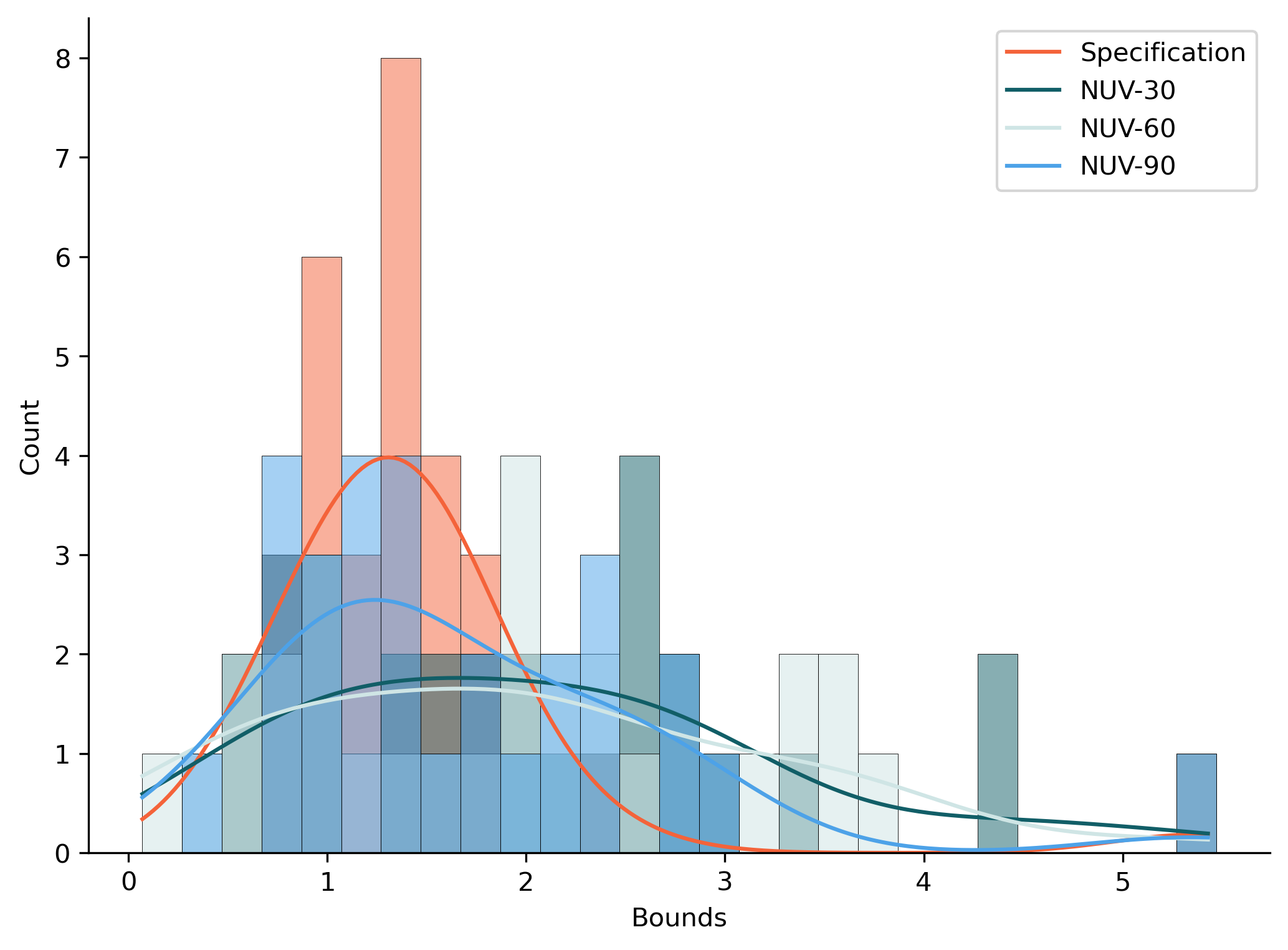}
    \caption{Illustrating the distribution of MAP upper bounds over 50 queries for the specification and NUV models.}
    \label{fig:its_bound_plot}
\end{figure}

Here, our NUVs simulate students interacting with the ITS. Specifically, in a typical use-case, K-12 students work on different topics sequentially. Therefore, to model this, we train the NUVs over a sample of problems over different topics. A NUV trained over $t\%$ topics is denoted by NUV-t.
The verification results are illustrated in the plot in Fig. \ref{fig:its_bound_plot}. Here, we show the distribution of the MAP upper bounds (the lower bounds follow a similar plot, but we do not show it for lack of space) for three NUV models over all hypercubes. As seen here, the NUV bounds approach the specification bounds as we cover more topics. That is, we are able to transfer the embeddings better when we cover more topics. In the real world, this implies that the ITS can provide better hints to students if the students interact with the ITS over greater number of topics.

\eat{
\begin{figure}
    \centering
    \includegraphics[scale=0.43]{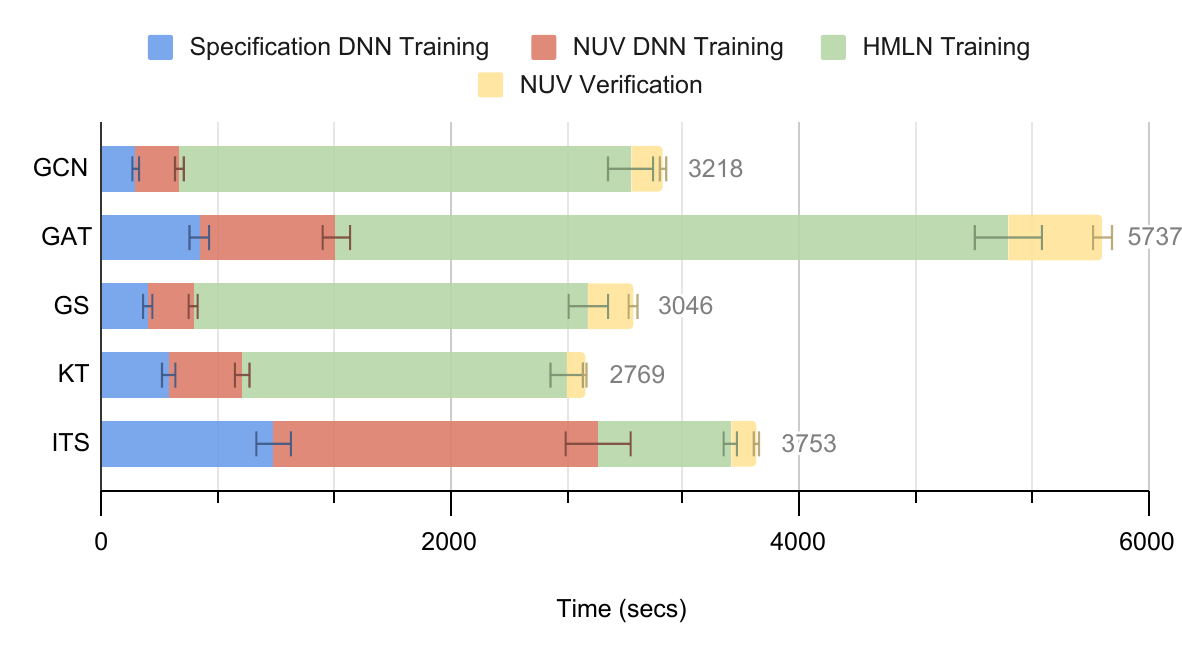}
    \caption{Cumulative time required for the entire verification process for the different DNNs.}
    \label{fig:verification_time}
\end{figure}

Finally, we show the computational requirements of our approach in Fig.~\ref{fig:verification_time}. Typically, training the specification HMLN was the most time-consuming, however, running verification is relatively inexpensive in our approach.
}



%% file: hmlns.tex
\begin{table}
    \centering
    \caption{Specification HMLNs for the three tasks. The property that we verify is shaded in each task. The predicate meanings are apparent from their names. The predicate ${\tt Dist}(e_{x},e_{y})$ denotes sub-symbolic atoms measuring distance between $e_{x},e_{y}$ which represent embeddings of $x$ and $y$.}
    \resizebox{0.485\textwidth}{!}{
    \tabcolsep=0.15cm
    \begin{tabular}{|c|c|}
        \hline
        \multirow{2}{*}{\makecell{\textbf{GNNs}}} & ${\tt Class}(x_1,c)$ $\wedge$ ${\tt Neighbor}(x_1,x_2)$ $\Rightarrow$ ${\tt Class}(x_2,c)$\\
         & \colorbox{gray!25}{${\tt Dist}(e_{x_1},e_{x_2})<\tau * ({\tt Class}(x_1,c)\Leftrightarrow{\tt Class}(x_2,c))$}\\
        \hline       
         \multirow{2}{*}{\textbf{KT}} & $({\tt Correct}(s,p_1)\wedge{\tt PreRequisite}(p_1,p_2)$\\
         &$\Rightarrow{\tt Correct}(s,p_2))$\\
        & \colorbox{gray!25}{${\tt Dist}(e_{s_1},e_{s_2})<\tau*({\tt Correct}(s_1,p)\Leftrightarrow{\tt Correct}(s_2,p))$}\\
         \hline
        \multirow{2}{*}{\textbf{ITS}} & ${\tt Success}(s,p_1)\wedge{\tt SameTopic}(p_1,p_2)\Rightarrow{\tt Success}(s,p_2)$\\         
        & \colorbox{gray!25}{${\tt Dist}(e_{s_1,p_1},e_{s_2,p_2})<\tau*({\tt Success}(s_1,p_1)\Leftrightarrow{\tt Success}(s_2,p_2))$}\\
        \hline
    \end{tabular}
    }
    \label{tab:hmlns}
\end{table}

%% file: results.tex
\begin{table*}
    \centering
    \caption{Verification results for GNNs. The t-statistic is shown in each case and the ones marked in red are those where the verification fails for the NUV, i.e., p-value $\leq$ $0.05$ for the Welch T-Test. UB refers to Upper bound and LB refers to Lower bound.}
    \resizebox{0.8\textwidth}{!}{
    \tabcolsep=0.15cm
    \begin{tabular}{|c|c|c|c|c|c|c|c|c|c|c|c|c|c|}
        \hline
        \multirow{3}{*}{\textbf{Spec.}} & \multirow{3}{*}{\textbf{NUV}} & \multicolumn{6}{|c|}{\textbf{Benchmarks}} & \multicolumn{6}{|c|}{\textbf{Noisy Benchmarks}}\\
        \cline{3-14}
        & &\multicolumn{2}{|c|}{\textbf{Cora}} & \multicolumn{2}{|c|}{\textbf{Citeseer}} & \multicolumn{2}{|c|}{\textbf{Pubmed}}&\multicolumn{2}{|c|}{\textbf{Cora}} & \multicolumn{2}{|c|}{\textbf{Citeseer}} & \multicolumn{2}{|c|}{\textbf{Pubmed}}\\
        \cline{3-14}
        & & \textbf{UB} & \textbf{LB} & \textbf{UB} & \textbf{LB} & \textbf{UB} & \textbf{LB} & \textbf{UB} & \textbf{LB} & \textbf{UB} & \textbf{LB} & \textbf{UB} & \textbf{LB}\\
         \hline
         \hline
         \multirow{2}{*}{GCN-S} 
         &GS& -0.692 & 1.21 & 0.427 & 1.33 & 0.447 & 1.27 & \textcolor{red}{2.227} & \textcolor{red}{2.71} & -1.22 & \textcolor{red}{2.66} & 0.6 & 0.276\\
         & GAT & -0.659 & 0.55 & 1.526 & 0.43 & 0.618 & -0.56 & -1.528 & -1.19 & 1.323 & 1.321 & 1.45 & \textcolor{red}{2.15}\\
         
         \hline
         
         \multirow{2}{*}{GS-S} & GCN & 0.395 & 1.83 & -0.55 & \textcolor{red}{2.722} & -0.58 & \textcolor{red}{2.29} & \textcolor{red}{-0.83} & -0.34 & -0.722 & \textcolor{red}{-0.623} & -0.45 & -0.29\\
             & GAT & \textcolor{red}{6.25} & 1.37 & \textcolor{red}{5.77} & 0.13 & 0.3 & -0.53 & \textcolor{red}{7.33} & \textcolor{red}{5.44} & 1.71 & \textcolor{red}{2.94} & 0.2 & 0.19\\
         
         \hline
         
         \multirow{2}{*}{GAT-S} & GCN & \textcolor{red}{2.019} & 0.56 & -1.53 & 1.21 & 1.8 & 1.28 & -1.65 & 0.33 & -2.053 & 2.042 & 1.8 & 1.81\\
         & GS & -0.91 & 0.772 & -1.56 & 1.56 & 1.763 & \textcolor{red}{2.13} & -0.96 & -1.15 & -1.61 & \textcolor{red}{-2.19} & 1.76 & 1.81\\
         \hline

    \end{tabular}
    }
    
    \label{tab:gnn-results}
\end{table*}

%% file: conclusion.tex
\section{Conclusion}
We presented a general approach for verification based on a first-order probabilistic model. Specifically, we encoded verifiable properties by relating symbolic domain knowledge with sub-symbolic DNN terms and parameterized these properties as a HMLN. To perform verification, we computed bounds using MILP solvers on the probabilities of a property. We illustrated verification in our framework using GNNs, Deep Knowledge Tracing and Intelligent Tutoring Systems.




%% file: main.bbl
\begin{thebibliography}{10}
\providecommand{\url}[1]{#1}
\csname url@samestyle\endcsname
\providecommand{\newblock}{\relax}
\providecommand{\bibinfo}[2]{#2}
\providecommand{\BIBentrySTDinterwordspacing}{\spaceskip=0pt\relax}
\providecommand{\BIBentryALTinterwordstretchfactor}{4}
\providecommand{\BIBentryALTinterwordspacing}{\spaceskip=\fontdimen2\font plus
\BIBentryALTinterwordstretchfactor\fontdimen3\font minus \fontdimen4\font\relax}
\providecommand{\BIBforeignlanguage}[2]{{%
\expandafter\ifx\csname l@#1\endcsname\relax
\typeout{** WARNING: IEEEtran.bst: No hyphenation pattern has been}%
\typeout{** loaded for the language `#1'. Using the pattern for}%
\typeout{** the default language instead.}%
\else
\language=\csname l@#1\endcsname
\fi
#2}}
\providecommand{\BIBdecl}{\relax}
\BIBdecl

\bibitem{embeddingcomparator}
A.~Boggust, B.~Carter, and A.~Satyanarayan, ``Embedding comparator: Visualizing differences in global structure and local neighborhoods via small multiples,'' in \emph{IUI}, 2022, p. 746–766.

\bibitem{wang2008hmln}
J.~Wang and P.~Domingos, ``Hybrid markov logic networks,'' in \emph{AAAI}, 2008, p. 1106–1111.

\bibitem{piech2015deep}
C.~Piech, J.~Bassen, J.~Huang, S.~Ganguli, M.~Sahami, L.~J. Guibas, and J.~Sohl-Dickstein, ``Deep knowledge tracing,'' in \emph{NeurIPS}, 2015, pp. 505--513.

\bibitem{albarghouthi-book}
A.~Albarghouthi, \emph{Introduction to Neural Network Verification}.\hskip 1em plus 0.5em minus 0.4em\relax verifieddeeplearning.com, 2021, \url{http://verifieddeeplearning.com}.

\bibitem{batten2021verification}
B.~Batten, P.~Kouvaros, A.~Lomuscio, and Y.~Zheng, ``Efficient neural network verification via layer-based semidefinite relaxations and linear cuts,'' in \emph{IJCAI}, 2021, pp. 2184--2190.

\bibitem{Katz19Marabou}
G.~Katz, D.~A. Huang, D.~Ibeling, K.~Julian, C.~Lazarus, R.~Lim, P.~Shah, S.~Thakoor, H.~Wu, A.~Zeljic, D.~L. Dill, M.~J. Kochenderfer, and C.~W. Barrett, ``The marabou framework for verification and analysis of deep neural networks,'' in \emph{CAV}, 2019, pp. 443--452.

\bibitem{ijcai2022p503}
X.~Xie, K.~Kersting, and D.~Neider, ``Neuro-symbolic verification of deep neural networks,'' in \emph{IJCAI}, 2022, pp. 3622--3628.

\bibitem{NEURIPS2019_5df0385c}
P.~Prabhakar and Z.~Rahimi~Afzal, ``Abstraction based output range analysis for neural networks,'' in \emph{NeurIPS}, 2019, pp. 15\,762--15\,772.

\bibitem{gagandeep2019}
G.~Singh, T.~Gehr, M.~P\"{u}schel, and M.~Vechev, ``An abstract domain for certifying neural networks,'' \emph{Proc. ACM Program. Lang.}, vol.~3, no. POPL, 2019.

\bibitem{bastani2016measuring}
O.~Bastani, Y.~Ioannou, L.~Lampropoulos, D.~Vytiniotis, A.~Nori, and A.~Criminisi, ``Measuring neural net robustness with constraints,'' \emph{NeurIPS}, vol.~29, 2016.

\bibitem{Tjeng19mip}
V.~Tjeng, K.~Y. Xiao, and R.~Tedrake, ``Evaluating robustness of neural networks with mixed integer programming,'' in \emph{ICLR}, 2019.

\bibitem{Botoeva2020}
E.~Botoeva, P.~Kouvaros, J.~Kronqvist, A.~Lomuscio, and R.~Misener, ``Efficient verification of relu-based neural networks via dependency analysis,'' \emph{AAAI 2020}, vol.~34, no.~04, pp. 3291--3299, 2020.

\bibitem{domingos&lowd09}
P.~Domingos and D.~Lowd, \emph{{Markov Logic: An Interface Layer for Artificial Intelligence}}.\hskip 1em plus 0.5em minus 0.4em\relax San Rafael, CA: Morgan \& Claypool, 2009.

\bibitem{Singla_Nath_Domingos_2014}
P.~Singla, A.~Nath, and P.~Domingos, ``Approximate lifting techniques for belief propagation,'' \emph{AAAI}, 2014.

\bibitem{Carkhel_Gogate_2018}
L.~Chou, P.~Sahoo, S.~Sarkhel, N.~Ruozzi, and V.~Gogate, ``Automatic parameter tying: A new approach for regularized parameter learning in markov networks,'' \emph{AAAI}, vol.~32, no.~1, 2018.

\bibitem{welch1947}
B.~L. Welch, ``The generalization of `student's' problem when several different population variances are involved,'' \emph{Biometrika}, pp. 28--35, 1947.

\bibitem{kipf2017semisupervised}
T.~N. Kipf and M.~Welling, ``Semi-supervised classification with graph convolutional networks,'' in \emph{ICLR}, 2017.

\bibitem{gat2018}
P.~Veli{\v{c}}kovi{\'{c}}, G.~Cucurull, A.~Casanova, A.~Romero, P.~Li{\`{o}}, and Y.~Bengio, ``{Graph Attention Networks},'' \emph{ICLR}, 2018.

\bibitem{hamilton2018inductive}
W.~L. Hamilton, Z.~Ying, and J.~Leskovec, ``Inductive representation learning on large graphs,'' in \emph{NeurIPS}, 2017, pp. 1024--1034.

\bibitem{pmlr-v206-alchihabi23a}
A.~Alchihabi and Y.~Guo, ``Learning robust graph neural networks with limited supervision,'' in \emph{AISTATS}, 2023, pp. 8723--8733.

\bibitem{irt1990}
F.~Drasgow and C.~L. Hulin, \emph{Item response theory}.\hskip 1em plus 0.5em minus 0.4em\relax Consulting Psychologists Press, 1990.

\bibitem{StamperKBSLDYS11a}
J.~C. Stamper, K.~R. Koedinger, R.~S.~J. de~Baker, A.~Skogsholm, B.~Leber, S.~Demi, S.~Yu, and D.~Spencer, ``Datashop: {A} data repository and analysis service for the learning science community,'' in \emph{{AIED}}, vol. 6738, 2011, p. 628.

\bibitem{shakya_23}
A.~Shakya, V.~Rus, and D.~Venugopal, ``{Scalable and Equitable Math Problem Solving Strategy Prediction in Big Educational Data},'' in \emph{EDM}, 2023, pp. 137--148.

\end{thebibliography}
